\documentclass{article}
\usepackage{log_2022}            

\usepackage{booktabs}            
\usepackage{multirow}            
\usepackage{amsfonts}            
\usepackage{graphicx}            
\usepackage{duckuments}          
\usepackage{amsmath}             

\usepackage[numbers,compress,sort]{natbib}

\title{
Early-Exit Graph Neural Networks for Link Prediction
}

\author{
Roman Knyazhitskiy \\
\institute{University of Cambridge} \\
\email{rk804@cam.ac.uk}
\And
Andrea Giuseppe Di Francesco \\
\institute{Sapienza University of Rome}
}

\begin{document}

\maketitle

\begin{abstract}
Graph Neural Networks are great for link prediction in various network-like structures; however, the question of their speed/quality tradeoff has been barely studied. While in practice the time it takes to do inference matters little for small benchmarks, the latency does limit applicability in large-scale domains. In this work, we explore early-exiting strategies that can be applied to Graph Neural Networks to solve the problem of link-prediction faster. We are largely inspired by the method described in \citep{early_exit_gnn}: we use no auxiliary losses to enforce early exiting, allowing it to emerge as an implicit property of the architecture. We show that our method enables early exiting in several setups, moving the Pareto frontier on the HeaRT benchmark for GCN and SAS-GNN backbones. Our findings show that inference speed of GNNs on many link-prediction problems can be improved, while losing little, or even winning in terms of prediction quality. The code is available in our repository: \url{https://github.com/knyazer/link_prediction}.
\end{abstract}

\section{Introduction}

Graph Neural Networks (GNNs) are the standard tool for learning on graph-structured data. Protein folding \citep{jumper2021highly}, citation networks \citep{Kipf:2017tc}, and social networks \citep{Hamilton:2017tp} are domains where the inductive biases of GNNs seem to work well.

A fundamental issue for deep GNN architectures is \emph{over-smoothing}: as depth increases, node embeddings converge to indistinguishable vectors \citep{nt2019, rusch23}. Attention-based GNNs mitigate this to some extent by being better at filtering irrelevant messages, but most architectures still saturate beyond roughly 20 layers \citep{rusch23}. Several remedies exist, including stochastic regularization \citep{zhang24}  and architectures with provable stability guarantees such as the Symmetric-Anti-Symmetric GNN (SAS-GNN) \citep{early_exit_gnn,rusch22}.

Beyond over-smoothing, GNNs are \emph{non-adaptive}: every node undergoes the same number of message-passing steps regardless of local difficulty. Consider a graph with two highly connected subgraphs - one dense (diameter $< 3$) and another sparse (diameter $> 5$). Predicting new links in the dense subgraph is inherently easier; spending the same compute on both wastes resources on the harder, less predictable subgraph.

\emph{Early exiting} \citet{scardapane2020} addresses this by detecting, via intermediate representations, whether further computation is warranted. Most early-exit methods require an auxiliary budget loss to encourage halting \citep{spinelli2021}. \citet{early_exit_gnn} demonstrated that for GNNs built on stable Neural ODE backbones, early exiting emerges naturally without such losses: later layers can have measurably worse prediction quality than earlier ones, so the task loss alone drives the exit policy.

Their work, however, targets node and graph classification. \emph{Link prediction} introduces additional challenges. First, predictions depend on \emph{pairs} of node embeddings, so freezing one node's representation affects all candidate edges involving that node. Second, the relevant subgraph structure differs across candidate edges, making per-edge adaptive compute especially appealing. To our knowledge, no prior work has applied auxiliary-loss-free early exiting to link prediction.

\begin{figure}[t]
	\centering
	\includegraphics[width=1\linewidth]{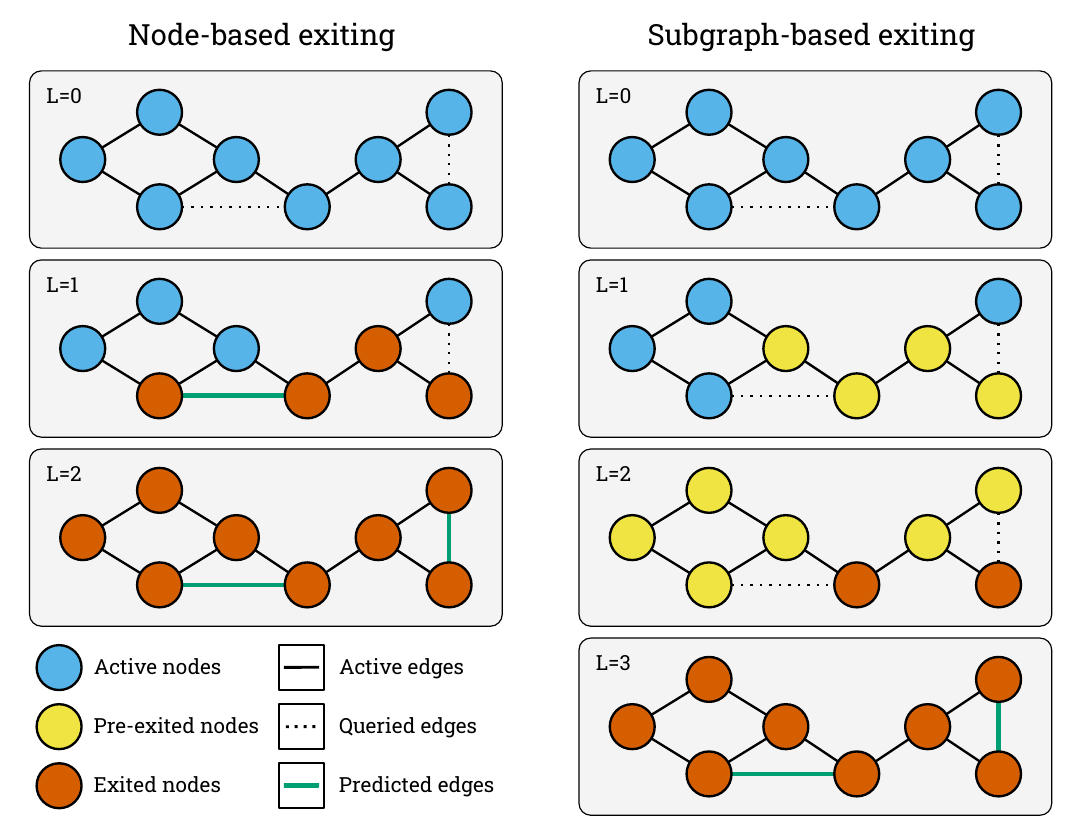}
	\caption{A visualization of both of our methods. \emph{Left:} node-based exiting, where the link is predicted when both of the nodes it depended on are exited. \emph{Right:} subgraph-based exiting, where the link is predicted when both of the nodes it depended on exited, but the neighbourhood around each node is also marked as "ready to exit".}
	\label{fig:cover-page}
\end{figure}

Our contributions are two-fold:
\begin{enumerate}
    \item We propose \textbf{node-level} and \textbf{subgraph-level} early-exit strategies for link prediction in GNNs. Both are trained end-to-end with only the task loss, following the EEGNN framework \citep{early_exit_gnn}, and we adapt them to operate on the HeaRT benchmark \citep{heart_bench}.
    \item We demonstrate empirically that early exiting shifts the performance-compute Pareto frontier on most HeaRT datasets for both GCN and SAS-GNN backbones, providing higher MRR and Hit@ metrics for the smaller compute budget.
\end{enumerate}

\section{Method}

\paragraph{Notation.}
Let $\mathcal{G} = (\mathcal{V}, \mathcal{E}, \mathbf{X})$ be an undirected graph with node set $\mathcal{V}$, edge set $\mathcal{E} \subseteq \mathcal{V} \times \mathcal{V}$, and node feature matrix $\mathbf{X} \in \mathbb{R}^{n \times d}$, where $n = |\mathcal{V}|$. The adjacency matrix is $\mathbf{A} \in \{0,1\}^{n \times n}$ and the normalised adjacency matrix is $\bar{\mathbf{A}} = \mathbf{D}^{-1/2}\mathbf{A}\mathbf{D}^{-1/2}$, where $\mathbf{D}$ is the degree matrix. At layer $l$, the hidden node embedding matrix is $\mathbf{H}^l \in \mathbb{R}^{n \times m}$, with $\mathbf{H}^0 = f_{\text{enc}}(\mathbf{X})$ for a learnable encoder $f_{\text{enc}}$. For a node $u$, we write $\mathbf{h}_u^l \in \mathbb{R}^m$ for its embedding at layer $l$.

\paragraph{Link prediction setting.}
Given a candidate edge $(u,v)$, we predict its existence from node embeddings $\mathbf{h}_u^l$ and $\mathbf{h}_v^l$ via a decoder $g: E \times E \to \mathbb{R}^2$.

\paragraph{Backbone.}
We experiment with two backbones: a standard GCN \citep{Kipf:2017tc}, with small tweaks from the HeaRT codebase, and SAS-GNN \citep{early_exit_gnn}. The SAS-GNN update rule is:
\begin{equation}
    \mathbf{H}^{l+1} = \mathbf{H}^l + \tau\,\sigma_1\!\bigl(-\sigma_2(\mathbf{H}^l \boldsymbol{\Omega}_{\text{as}}) + \bar{\mathbf{A}}\mathbf{H}^l \mathbf{W}_s\bigr),
    \label{eq:sas_update}
\end{equation}
where $\boldsymbol{\Omega}_{\text{as}} = \boldsymbol{\Omega} - \boldsymbol{\Omega}^\top$ is an antisymmetric weight matrix, $\mathbf{W}_s = \mathbf{W}_s^\top$ is a symmetric weight matrix, $\tau$ is the integration step, and $\sigma_1 = \mathrm{ReLU} \circ \tanh$, $\sigma_2 = \mathrm{ReLU}$. This ODE-inspired design is provably stable and non-dissipative \citep{early_exit_gnn, Hamilton:2017tp}, ensuring that intermediate representations remain informative. While originally argued to be a prerequisite for early-exiting, we find in practice that early exiting is feasible for standard GNN architectures too. We show this in Section 4.

\paragraph{Gumbel-Softmax exit mechanism.}
Following \citet{early_exit_gnn}, we attach a lightweight confidence head at each node and each layer. For node $u$ at layer $l$, a shared GNN module $f_c$ produces a confidence vector $\mathbf{C}_u^l \in \mathbb{R}^2$ over actions $\{$\textsc{continue}, \textsc{exit}$\}$. A second module $f_\nu$ predicts a temperature $\nu_u^l > 0$. The Gumbel-Softmax score is:
\begin{equation}
    \mathbf{c}_u^l = \mathrm{Softmax}\!\left(\frac{\log \mathbf{C}_u^l + \mathbf{g}_u^l}{\nu_u^l}\right), \quad \mathbf{g}_u^l \sim \mathrm{Gumbel}(0,1),
    \label{eq:gumbel}
\end{equation}
which approaches a one-hot vector as $\nu_u^l \to 0$. We use the straight-through estimator \citep{jang2017} for backpropagation. The hidden dimension of $f_c$ and $f_\nu$ is usually chosen to be $m_f = 32$, compared to $m = 256$ for node embeddings, keeping overhead small.

The reason to choose Gumbel is purely heuristic: it seems to behave better in our setup than the standard categorical straight-through estimator.

\subsection{Node-based early exiting}

\begin{figure}[t]
	\centering
	\includegraphics[width=1\linewidth]{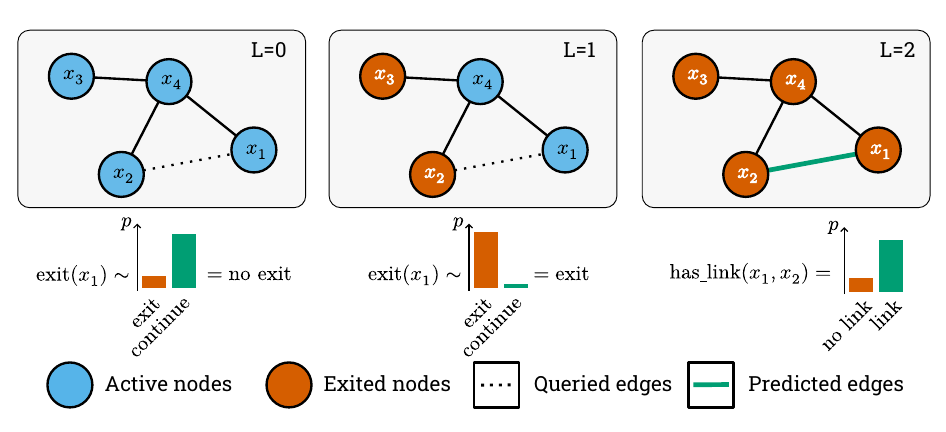}
	\caption{Node based early exiting visualization. On each layer, each node decides whether it wants to exit - which implies that the embeddings of the node will be frozen. It does so by doing a relaxation between greedy and categorical sampling from its decision (continue/exit) output logits with gumbel-softmax trick.}
	\label{fig:node-based-exiting}
\end{figure}

Each node independently decides whether to exit at each layer. The integration step in Eq. \eqref{eq:sas_update} becomes node- and layer-dependent:
\begin{equation}
    \tau_u^l = c_u^l(0),
    \label{eq:tau_node}
\end{equation}
where $xc_u^l(0)$ is the \textsc{continue} probability. When a node exits ($c_u^l(0) \to 0$), its embedding is frozen: $\mathbf{h}_u^{l+1} = \mathbf{h}_u^l$, but the node \emph{still participates in message passing} for its neighbours' updates. The final embedding used for link prediction is $\mathbf{z}_u = \mathbf{h}_u^{l^*}$, where $l^*_u = \min\{l : \arg\max \mathbf{c}_u^l = 1\}$, or $l^* = L$ if the node never exits.

The heads $f_c$ and $f_\nu$ are trained with only the link prediction loss (e.g. binary cross-entropy over positive and negative edges). No auxiliary budget loss is used; early exiting emerges from the task objective alone.

\subsection{Subgraph-based early exiting}

  Node-level exiting might lead to some inconsistencies: a node may freeze while the neighbors still update, which might create a mismatch in the information flow of adjacent embeddings; for instance, maybe a node is aware of the need to propagate its embedding to neighbors, but neighbors are not aware of that and just exit. To address this, we introduce an exit protocol that conditions each
   node's exit on its neighborhood's readiness.

  \paragraph{Phase 1: Soft exit.}
  At each layer $l$, a shared confidence head $f_s$ predicts a soft exit score for each node. Using the Gumbel-Softmax mechanism from Eq.~\eqref{eq:gumbel}, we obtain a binary signal $e_u^l \in \{0,1\}$ indicating whether node $u$ considers itself ready to
   exit. We track the cumulative soft-exit state $\bar{e}_u^l = 1 - \prod_{k=1}^{l}(1 - e_u^k)$, which increases as the node becomes more confident.

  \paragraph{Phase 2: Hard exit via neighborhood readiness.}
  Before a node actually freezes, we check whether its local neighborhood shares its confidence. We compute a readiness score as the mean soft-exit state of its neighbors:
  \begin{equation}
      R_u^l = \frac{1}{|\mathcal{N}(u)|} \sum_{v \in \mathcal{N}(u)} \bar{e}_v^l.
      \label{eq:readiness}
  \end{equation}
  A separate hard exit head $f_h$, a two-layer MLP, takes the concatenation $[\mathbf{h}_u^l;\, R_u^l]$ as input and produces logits over $\{\textsc{continue}, \textsc{exit}\}$, which are again passed through Gumbel-Softmax. The effective exit probability
  is gated by the node's own soft-exit state, so that a node can only hard-exit once it has marked itself as ready:
  \begin{equation}
      p_u^l = \bar{e}_u^l \cdot c_u^{l,\text{hard}}(1),
      \label{eq:hard_exit}
  \end{equation}
  where $c_u^{l,\text{hard}}(1)$ is the hard exit probability, outputted by $f_h$. This forces dense subgraphs to exit together: a node waits until its neighbors are also ready before freezing its embedding.

\subsection{Theoretical Compute. } \label{sec:compute}
  
  We approximate the compute cost at layer $l$ as:
  \begin{equation}
      C^{(l)} = n^{(l)}_{\text{active}} \cdot h^2 + e^{(l)}_{\text{active}} \cdot h,
      \label{eq:compute}
  \end{equation}
  where $h$ is the hidden dimension. For the baseline (no early exiting), $n^{(l)}_{\text{active}} = N$ and $e^{(l)}_{\text{active}} = E$ at every layer. Note that these FLOP savings should correspond to wall-clock speedups on CPU; on GPU, realising them requires sparse-execution support or custom kernels, since frozen nodes still occupy tensor rows in a naive dense implementation.
  
\subsection{Implementation Notes}
\label{sec:impl_notes}

We largely follow the EEGNN implementation of \citep{early_exit_gnn}. We identified one critical instability: the original temperature function $f_\nu$ can output values near zero, causing the softmax gradients in Eq. \eqref{eq:gumbel} to diverge to infinity or produce NaN values if trained for too long. We resolve this by clamping $\nu_u^l$ from below:
\begin{equation}
    \nu_u^l = \max\!\bigl(f_\nu(\mathbf{H}^l),\; 0.1\bigr).
    \label{eq:temp_clamp}
\end{equation}
This simple fix stabilises training for deep networks ($L \geq 20$) and is necessary for convergence. The threshold $0.1$ was not tuned; moderate variations (e.g.\ $0.05$--$0.2$) did not noticeably affect performance.

\begin{figure}[h!]
	\centering
	\includegraphics[width=1.0\linewidth]{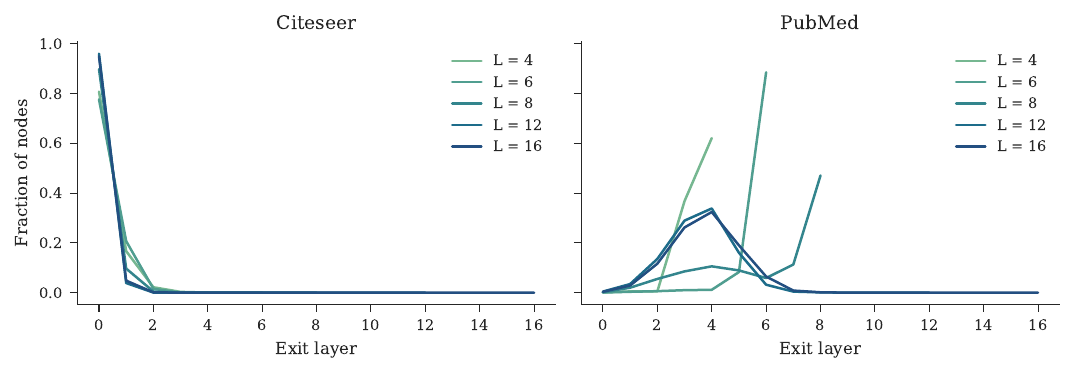}
	\caption{Early exiting depth distribution of GCN for PubMed and Citeseer. While Citeseer seems to be an easy problem and collapses to the very small depth, PubMed behaves more interestingly. While for smaller depths it learns to never exit, it shows a distinct bump around layer 4 for deep networks; most of the nodes exit around this layer, with few exiting before and after. Shown are the averages of three training runs.}
	\label{fig:exit_probability_per_layers}
\end{figure}

\section{Evaluation}

\subsection{Early exit depth distribution}

We look at how the early exit depth is distributed for a few datasets; as in, does the exit depth distribution change meaningfully for deeper nets. We notice an interesting fact: while for deeper networks there is an emergence of the desired behaviour, it appears as a consistent bump near the intrinsic complexity of the dataset.
Figure~\ref{fig:exit_probability_per_layers} shows the exit layer distribution for ResGCN on Citeseer and PubMed across depths $L \in \{4, 6, 8, 12, 16\}$. On Citeseer, the distribution is sharply concentrated at the first few layers regardless of network depth: nearly all nodes exit immediately, suggesting that the link prediction task on this graph requires very little message passing. This is consistent with Citeseer being a small, relatively dense citation network where local features are already highly informative; indeed, an MLP on node features alone can achieve competitive accuracy on this dataset. Surprisingly we can see two distinct modes: shallower networks are more reluctant to start exiting early, and have around $10\%$ probability of exiting at layer 2, while deeper networks collapse harder.

\subsection{Performance and Metrics}

\begin{figure}[h!]
	\centering
	\includegraphics[width=1\linewidth]{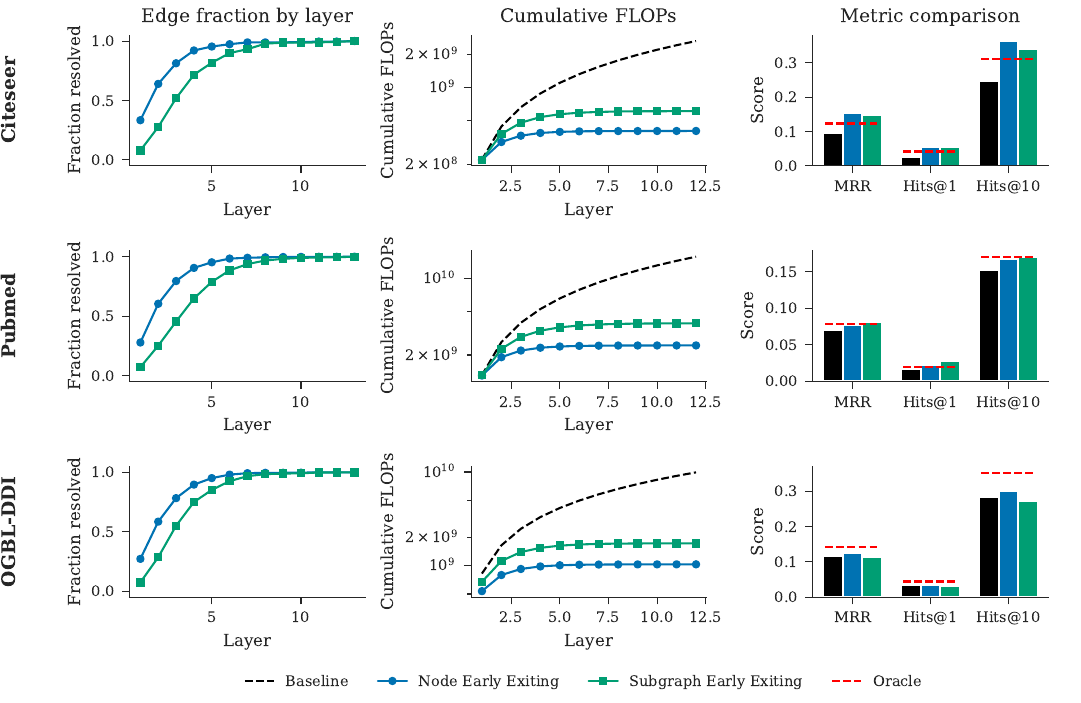}
	\caption{Evaluation of GCN via HeaRT on a few datasets. Our method outperforms the baseline, and sometimes even outperforms the oracle, which is a GCN with optimally chosen depth. Edge fraction by layer corresponds to the number of exited edges per layer, with 1 corresponding to all edges being done. Cumulative FLOPS are the amount of compute the model has spent up to a particular layer. }
	\label{fig:big_results_gcn}
\end{figure}

We evaluate node-level and subgraph-level early exiting on three HeaRT benchmark datasets: Citeseer, PubMed, and ogbl-ddi. We compare against a fixed-depth baseline (no early exiting) and an oracle that trains separate fixed-depth GCNs for $L \in \{1, 3, 5, 7, 9, 11\}$ and selects the depth with the highest validation MRR. All experiments use a GCN backbone with L=15 layers unless stated otherwise.

Figure~\ref{fig:big_results_gcn} summarises the results. The left column shows the fraction of nodes resolved (exited) as a function of layer. Node-level exiting resolves nodes faster than subgraph exiting on all three datasets, which is expected: the neighbourhood consensus constraint in subgraph exiting delays individual nodes from freezing. On Citeseer and PubMed, both methods resolve nearly all nodes by layer 8--10, well before the final layer.
The middle column shows cumulative FLOPs (computed via Eq. \ref{eq:compute}). Both early-exit strategies use substantially less compute than the baseline across all datasets. Node exiting is consistently cheaper than subgraph exiting, since it freezes nodes more aggressively. On Citeseer, node exiting uses roughly $5\times$ fewer FLOPs than the baseline at layer 15.

The right column shows link prediction quality (MRR, Hits@1, Hits@10). On Citeseer, subgraph exiting matches or exceeds the oracle on Hits@10, and both methods substantially outperform the baseline on all metrics. On PubMed, the picture is more mixed: node exiting underperforms the baseline on MRR and Hits@1, while subgraph exiting stays competitive. On ogbl-ddi, both methods improve over the baseline on Hits@10, though MRR results are comparable.

\subsection{Hyperparameter sweep}

For GCN evaluations we do 3 runs to tune the learning rate, and then 3 runs to compute the distribution of losses at a particular value, thus a total of 5 runs, with one being reused between the two stages. This is similar to a HeaRT benchmark setup, but with a lot of compute saved. We couldn't run some datasets due to them being too large or not well-supported by HeaRT implementation, and we decreased the maximum number of allowed steps by a factor of 10 to accomodate limited time. The results can be seen in Table \ref{tab:benchmark}.

\begin{table}[h!]
\centering
\caption{MRR (\%) on HeaRT benchmark (500 hard negatives). \textbf{Fixed}: standard GCN without early exiting. \textbf{NodeExit}: node-level early exiting (Section~2.1). \textbf{SubgraphExit}: subgraph-level early exiting (Section~2.2). LR tuned from \{0.0005, 0.001, 0.002\}, 3 seeds. Best result per row in \textbf{bold}.}
\label{tab:benchmark}
\begin{tabular}{ll ccc}
\toprule
Dataset & $L$ & Fixed & NodeExit & SubgraphExit \\
\midrule
\multirow{4}{*}{Citeseer} & 1 & $20.39{\scriptstyle \pm 1.03}$ & $\mathbf{20.51}{\scriptstyle \pm 1.15}$ & $20.30{\scriptstyle \pm 0.28}$ \\
 & 2 & $19.86{\scriptstyle \pm 0.68}$ & $\mathbf{20.94}{\scriptstyle \pm 0.67}$ & $20.24{\scriptstyle \pm 0.54}$ \\
 & 4 & $18.50{\scriptstyle \pm 0.53}$ & $\mathbf{22.82}{\scriptstyle \pm 0.53}$ & $19.53{\scriptstyle \pm 0.44}$ \\
 & 8 & $16.47{\scriptstyle \pm 0.11}$ & $\mathbf{24.42}{\scriptstyle \pm 0.62}$ & $22.64{\scriptstyle \pm 0.20}$ \\
\midrule
\multirow{4}{*}{PubMed} & 1 & $\mathbf{8.48}{\scriptstyle \pm 0.12}$ & $8.07{\scriptstyle \pm 0.11}$ & $8.31{\scriptstyle \pm 0.15}$ \\
 & 2 & $8.84{\scriptstyle \pm 0.16}$ & $\mathbf{9.19}{\scriptstyle \pm 0.11}$ & $9.09{\scriptstyle \pm 0.19}$ \\
 & 4 & $10.07{\scriptstyle \pm 0.14}$ & $9.66{\scriptstyle \pm 0.12}$ & $\mathbf{10.13}{\scriptstyle \pm 0.20}$ \\
 & 8 & $9.61{\scriptstyle \pm 0.21}$ & $\mathbf{9.65}{\scriptstyle \pm 0.15}$ & $9.36{\scriptstyle \pm 0.30}$ \\
\midrule
\multirow{4}{*}{ogbl-ddi} & 1 & $12.78{\scriptstyle \pm 0.97}$ & $13.75{\scriptstyle \pm 0.75}$ & $\mathbf{14.15}{\scriptstyle \pm 0.23}$ \\
 & 2 & $13.23{\scriptstyle \pm 0.63}$ & $13.23{\scriptstyle \pm 0.20}$ & $\mathbf{13.51}{\scriptstyle \pm 0.09}$ \\
 & 4 & $13.26{\scriptstyle \pm 0.22}$ & $\mathbf{13.76}{\scriptstyle \pm 0.38}$ & $12.99{\scriptstyle \pm 0.14}$ \\
 & 8 & $11.78{\scriptstyle \pm 0.42}$ & $\mathbf{13.99}{\scriptstyle \pm 0.34}$ & $13.96{\scriptstyle \pm 0.49}$ \\
\bottomrule
\end{tabular}
\end{table}

\section{Discussion and Conclusion}

Our method seems to work fairly well, offering a way to prevent over-smoothing while saving compute. However, we discovered that likely due to the non-enforcing nature of the algorithm, the behaviour of the method is very sensitive to hyperparameters and small implementation choices. The main issue is not that hyperparameters affect performance, but that the effect is hard to predict and rather stochastic: e.g. choosing a significantly wrong learning rate leads to our method collapsing into either instant exiting or never-exiting strategies. Or, for instance, when training a shallow GCN, it looks convergent around 500 steps to exiting very early; however, if you train it for 10\_000 more steps, it will change the behaviour to exiting at the last layer! Thus the early-exit distribution dynamics throughout training warrants closer inspection.

We noticed that the subgraph method reduces these effects significantly. By enforcing spatial consensus, it acts as a structural regulariser; in fact, the subgraph method is even trainable with a hard Gumbel estimator, instead of the soft one. Thus, we believe that the subgraph method, or similar neighbourhood-aware schemes, represent the more robust path forward for real-world applications.

\paragraph{Limitations.} Our evaluation is limited to three HeaRT datasets with training budgets reduced by $10\times$ relative to the standard protocol, owing to compute constraints. We report SAS-GNN results only in the Appendix for the same reason. The FLOP savings may not translate to GPU speedups without sparse-execution support (see Section~\ref{sec:compute}). Finally, we evaluate only two backbone architectures (GCN and SAS-GNN); generalisation to attention-based or heterogeneous GNNs remains untested.

We hope future work could address the lack of heavy explorations about different emergent behaviours of our methods.

\bibliographystyle{unsrtnat}
\bibliography{reference}

@inproceedings{Kipf:2017tc,
author = {Kipf, Thomas N. and Welling, Max},
title = {{Semi-Supervised Classification with Graph Convolutional Networks}},
booktitle = {ICLR},
year = {2017}
}

@inproceedings{Hamilton:2017tp,
author = {Hamilton, William L. and Ying, Zhitao and Leskovec, Jure},
title = {{Inductive Representation Learning on Large Graphs}},
booktitle = {NIPS},
year = {2017},
pages = {1024--1034},
}

@article{early_exit_gnn,
  title={Early-Exit Graph Neural Networks},
  author={Andrea Giuseppe Di Francesco and Maria Sofia Bucarelli and Franco Maria Nardini and Raffaele Perego and Nicola Tonellotto and Fabrizio Silvestri},
  journal={ArXiv},
  year={2025},
  volume={abs/2505.18088},
  url={https://api.semanticscholar.org/CorpusID:278886684}
}

@inproceedings{heart_bench,
 author = {Li, Juanhui and Shomer, Harry and Mao, Haitao and Zeng, Shenglai and Ma, Yao and Shah, Neil and Tang, Jiliang and Yin, Dawei},
 booktitle = {Advances in Neural Information Processing Systems},
 editor = {A. Oh and T. Naumann and A. Globerson and K. Saenko and M. Hardt and S. Levine},
 pages = {3853--3866},
 publisher = {Curran Associates, Inc.},
 title = {Evaluating Graph Neural Networks for Link Prediction: Current Pitfalls and New Benchmarking},
 url = {https://proceedings.neurips.cc/paper_files/paper/2023/file/0be50b4590f1c5fdf4c8feddd63c4f67-Paper-Datasets_and_Benchmarks.pdf},
 volume = {36},
 year = {2023}
}

@article{jumper2021highly,
  author  = {Jumper, John and Evans, Richard and Pritzel, Alexander and Green, Tim and Figurnov, Michael and Ronneberger, Olaf and Tunyasuvunakool, Kathryn and Bates, Russ and Žídek, Augustin and Potapenko, Anna and Bridgland, Alex and Meyer, Clemens and Kohl, Simon A. A. and Ballard, Andrew J. and Cowie, Andrew and Romera-Paredes, Bernardino and Nikolov, Stanislav and Jain, Rishub and Adler, Jonas and Back, Trevor and Petersen, Stig and Reiman, David and Clancy, Ellen and Zielinski, Michal and Steinegger, Martin and Pacholska, Michalina and Berghammer, Tamas and Bodenstein, Sebastian and Silver, David and Vinyals, Oriol and Senior, Andrew W. and Kavukcuoglu, Koray and Kohli, Pushmeet and Hassabis, Demis},
  title   = {Highly accurate protein structure prediction with AlphaFold},
  journal = {Nature},
  year    = {2021},
  volume  = {596},
  number  = {7873},
  pages   = {583--589},
  month   = aug,
  doi     = {10.1038/s41586-021-03819-2},
  url     = {https://doi.org/10.1038/s41586-021-03819-2},
  issn    = {1476-4687}
}

@article{scardapane2020,
  author   = {Scardapane, Simone and Scarpiniti, Michele and Baccarelli, Enzo and Uncini, Aurelio},
  title    = {Why Should We Add Early Exits to Neural Networks?},
  journal  = {Cognitive Computation},
  year     = {2020},
  volume   = {12},
  number   = {5},
  pages    = {954--966},
  month    = sep,
  doi      = {10.1007/s12559-020-09734-4},
  url      = {https://doi.org/10.1007/s12559-020-09734-4},
  issn     = {1866-9964}
}

@article{spinelli2021,
  author   = {Spinelli, Indro and Scardapane, Simone and Uncini, Aurelio},
  title    = {Adaptive Propagation Graph Convolutional Network},
  journal  = {IEEE Transactions on Neural Networks and Learning Systems},
  year     = {2021},
  volume   = {32},
  number   = {10},
  pages    = {4755--4760},
  month    = oct,
  doi      = {10.1109/TNNLS.2020.3025110},
  url      = {https://doi.org/10.1109/TNNLS.2020.3025110},
  issn     = {2162-2388}
}

@InProceedings{rusch22,
  title = 	 {Graph-Coupled Oscillator Networks},
  author =       {Rusch, T. Konstantin and Chamberlain, Ben and Rowbottom, James and Mishra, Siddhartha and Bronstein, Michael},
  booktitle = 	 {Proceedings of the 39th International Conference on Machine Learning},
  pages = 	 {18888--18909},
  year = 	 {2022},
  editor = 	 {Chaudhuri, Kamalika and Jegelka, Stefanie and Song, Le and Szepesvari, Csaba and Niu, Gang and Sabato, Sivan},
  volume = 	 {162},
  series = 	 {Proceedings of Machine Learning Research},
  month = 	 {17--23 Jul},
  publisher =    {PMLR},
  url = 	 {https://proceedings.mlr.press/v162/rusch22a.html}
}

@misc{rusch23,
      title={A Survey on Oversmoothing in Graph Neural Networks}, 
      author={T. Konstantin Rusch and Michael M. Bronstein and Siddhartha Mishra},
      year={2023},
      eprint={2303.10993},
      archivePrefix={arXiv},
      primaryClass={cs.LG},
      url={https://arxiv.org/abs/2303.10993}, 
}

@misc{nt2019,
      title={Revisiting Graph Neural Networks: All We Have is Low-Pass Filters}, 
      author={Hoang NT and Takanori Maehara},
      year={2019},
      eprint={1905.09550},
      archivePrefix={arXiv},
      primaryClass={stat.ML},
      url={https://arxiv.org/abs/1905.09550}, 
}

@article{zhang24,
   title={SSFG: Stochastically Scaling Features and Gradients for Regularizing Graph Convolutional Networks},
   volume={35},
   ISSN={2162-2388},
   url={http://dx.doi.org/10.1109/TNNLS.2022.3188888},
   DOI={10.1109/tnnls.2022.3188888},
   number={2},
   journal={IEEE Transactions on Neural Networks and Learning Systems},
   publisher={Institute of Electrical and Electronics Engineers (IEEE)},
   author={Zhang, Haimin and Xu, Min and Zhang, Guoqiang and Niwa, Kenta},
   year={2024},
   month=feb, pages={2223–2234} }

@misc{jang2017,
      title={Categorical Reparameterization with Gumbel-Softmax}, 
      author={Eric Jang and Shixiang Gu and Ben Poole},
      year={2017},
      eprint={1611.01144},
      archivePrefix={arXiv},
      primaryClass={stat.ML},
      url={https://arxiv.org/abs/1611.01144}, 
}

\appendix

\section{SAS exit plot}

\begin{figure}[h!]
	\centering
	\includegraphics[width=1\linewidth]{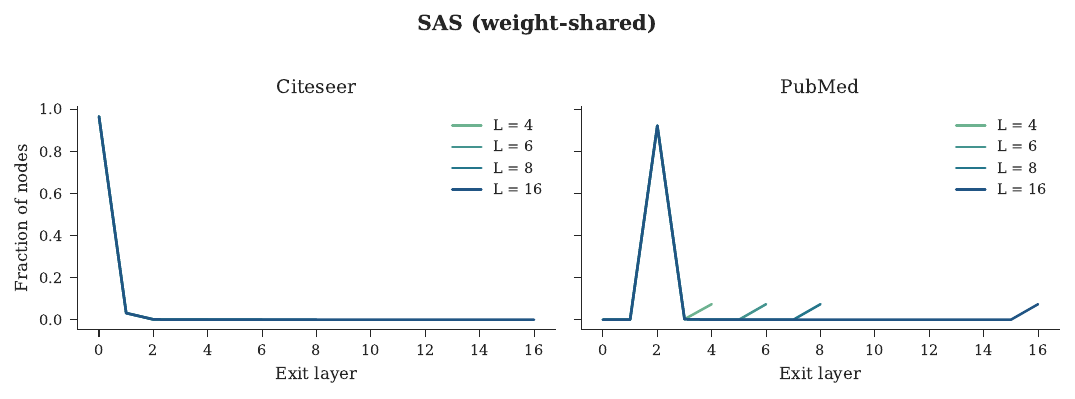}
	\caption{Early exiting depth distribution of SAS for PubMed and Citeseer. SAS seems to be a lot more prone to undesirable collapse than GCN, but also a lot more consistent.}
	\label{fig:sas_exit_probability_per_layers}
\end{figure}

\section{SAS results plot}

\begin{figure}[h!]
	\centering
	\includegraphics[width=1\linewidth]{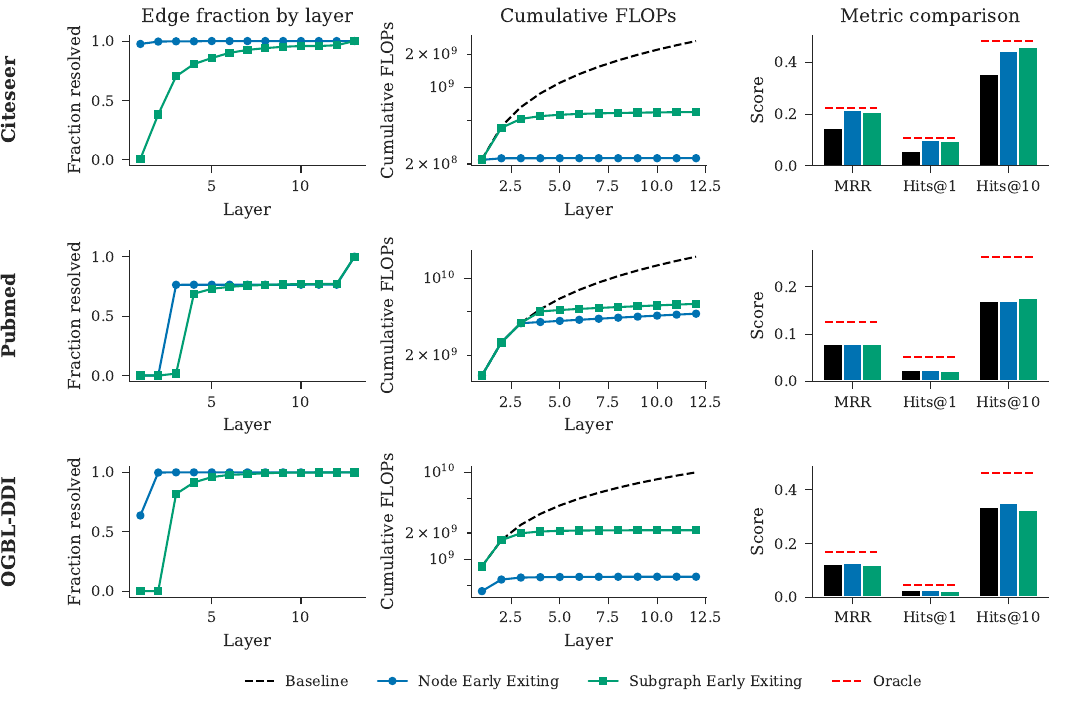}
	\caption{Evaluation of SAS via HeaRT on a few datasets. Our method outperforms the baseline, but does not reach the oracle performance. }
	\label{fig:big_results_sas}
\end{figure}

\end{document}